\documentclass[a4paper, 10pt, conference]{ieeeconf}      

\IEEEoverridecommandlockouts                              

\usepackage{epsfig} 
\usepackage{mathptmx} 
\usepackage{times} 
\usepackage{amsmath} 
\usepackage{amssymb}  
\usepackage{graphicx}
\usepackage{booktabs}
\usepackage{color}
\usepackage{multirow}
\usepackage{makecell}
\usepackage{caption} 
\usepackage{subcaption}

\title{\LARGE \bf
PNeRF: Probabilistic Neural Scene Representations \\ for Uncertain 3D Visual Mapping
}

\author{Yassine Ahmine$^{1}$, Arnab Dey$^{1}$ and Andrew I. Comport$^{1}$
\thanks{$^{1}$I3S-CNRS/Université Cote d’Azur
Sophia-Antipolis, France}%
}

\begin{document}

\maketitle

\begin{abstract}

Recently neural scene representations have provided very impressive results for representing 3D scenes visually, however, their study and progress have mainly been limited to visualization of virtual models in computer graphics or scene reconstruction in computer vision without explicitly accounting for sensor and pose uncertainty. Using this novel scene representation in robotics applications, however, would require accounting for this uncertainty in the neural map. The aim of this paper is therefore to propose a novel method for training {\em probabilistic neural scene representations} with uncertain training data that could enable the inclusion of these representations in robotics applications. Acquiring images using cameras or depth sensors contains inherent uncertainty, and furthermore, the camera poses used for learning a 3D model are also imperfect. If these measurements are used for training without accounting for their uncertainty, then the resulting models are non-optimal, and the resulting scene representations are likely to contain artifacts such as blur and un-even geometry. In this work, the problem of uncertainty integration to the learning process is investigated by focusing on training with uncertain information in a probabilistic manner. The proposed method involves explicitly augmenting the training likelihood with an uncertainty term such that the learnt probability distribution of the network is minimized with respect to the training uncertainty. It will be shown that this leads to more accurate image rendering quality, in addition to more precise and consistent geometry. Validation has been carried out on both synthetic and real datasets showing that the proposed approach outperforms state-of-the-art methods. The results show notably that the proposed method is capable of rendering novel high-quality views even when the training data is limited.

\end{abstract}

\section{Introduction}

Probabilistic representations are at the heart of robotics~\cite{thrun2005probabilistic} and allow for a wide range of applications, including SLAM and visual odometry, by providing formal methods to account for uncertain sensor data when measuring and representing the environment. 3D scene representations using multilayer perceptron neural networks (MLP), have witnessed overwhelming interest from neighboring scientific communities during the last few years, notably because of the impressive level of realism achieved in representing scenes in this way. However, these methods do not explicitly account for uncertainties in their representations. Sensors measuring real-world quantities inherently contain some uncertainty (such as cameras, depth sensors, etc.), and this should be accounted for in the case where a robot interacts with its environment. Subsequently, the aim of this article is to investigate the problem of incorporating uncertainty into neural scene representations and to show its importance in terms of geometric consistency and increased map quality. 

In robotics, 3D scene representations have often been modeled in various ways, including voxels~\cite{newcombe2011dtam, whelan2013vo} and key-frame graphs~\cite{meilland2011vslam, meilland2013unifying}, which aim to model scene surfaces. In this case, different primitives can be considered, for example, point clouds \cite{keller2013Realtime, wiles19synsin}, meshes \cite{nimier19mitsuba}, or implicit surfaces \cite{curless96avolumetric, takikawa2021nglod}. In ADOP (Approximate Differentiable One-Pixel Point Rendering) \cite{ruckert2021adop}, for instance, 3D point clouds are used to store learnable features that are interpreted by an MLP (Multilayer Perceptron) to produce view-dependent effects. Furthermore, Deferred Neural Rendering \cite{thies19DNR} represent surfaces as meshes, proposing to use the so-called neural textures as feature maps for the meshes. In addition to that, \cite{zhang2021learning} uses a signed distance field and a surface light field to represent scenes, taking advantage of a depth estimator network to supervise its geometry branch. Surface-based representation only models the information present on the surface of the scene elements, allowing this method to limit its memory footprint. However, this comes at the cost of not being able to accurately model volumetric matter \cite{tewari2021advances}. 

Volumetric scene representations are more general, as they permit to model solid elements such as surfaces along with volumetric matter as smoke. Historically, voxel grids were the first volumetric approaches to be developed \cite{connolly84octree, chien88octree}. They offer a well-defined formalism for modeling 3D scenes. However, such approaches suffer from important memory footprints, which can become prohibitively large when the image resolution is high \cite{tewari2021advances}. 

The past years have witnessed the development of a new kind of approach for volumetric scene representation that combines methods from \textit{machine learning} with those from \textit{computer vision} and \textit{computer graphics}, known as neural scene representation. An example of such approaches is Neural Volumes \cite{lombardi19NV}, which proposes to train a 3D CNN (Convolutional Neural Network) to reconstruct single scenes from multiview images. Another kind of approach that allowed to produce photorealistic images using classical image rendering is NeRF (Neural Radiance Fields) \cite{mildenhall2020nerf}. This approach is able to take into account specularity effects, while being able to model high-frequency image elements due to the use of positional encoding. It also exhibits interesting compression properties, which come at the cost of intensive computations, as several seconds are required to render a single high-resolution image on a typical desktop GPU. Consequently, different methods were proposed to overcome this limitation \cite{yu2021plenoctrees, yu2021pixelnerf} and improve the performance of the approach \cite{zhang2020nerf, mueller2022instant}. In addition, other approaches focused on the use of depth information to improve the modeled geometry and the quality of the rendered image, such as \cite{wei2021nerfingmvs, neff2021donerf, dey2022mip}.

The aforementioned approaches related to NeRF make the assumption that the camera poses for the training images are known and rely on SFM (Structure From Motion) methods such as COLMAP \cite{schoenberger2016mvs,schoenberger2016sfm} to produce the desired poses in the case of real data. However, some approaches have been proposed to estimate camera poses along with neural network parameters \cite{SCNeRF2021,wang2021nerfmm,lin2021barf}. Therefore, self-calibrating neural radiation fields \cite{SCNeRF2021} proposes to additionally estimate intrinsic, extrinsic, and distortion camera parameters; while BARF \cite{lin2021barf} proposes to apply an annealing schedule to each component of the positional encoding for coarse-to-fine trajectory estimation.

Other works aimed at using neural radiance fields in robotics have recently been developed. For instance, \cite{sucar2021imap} and \cite{Zhu2022nice} proposed SLAM systems that use NeRF-based map representation. Other approaches such as \cite{yen2020inerf} aimed to solve the inverse problem of estimating the pose of a query image using a trained NeRF. Furthermore, \cite{adam2022ral} used the output of NeRF to build a trajectory planning algorithm, and~\cite{yen2022nerfsupervision} used this kind of approach to learn view invariant object descriptors. These approaches, however, didn't take explicitly model the uncertainty distribution, which does not allow them to best exploit the available information.  

In this work, the objective is integrate such an uncertainty distribution into the learning process of NeRF, which is beneficial for various robotics applications such as SLAM. Therefore, the learned distribution of depths along view rays is supervised using the available geometric information, such as point cloud estimates or depth measurements. This supervision during the learning process of neural radiance fields, in addition to being an inductive bias towards the correct scene geometry, allows a direct integration of uncertainty; resulting in an increased rendering quality and better geometry modeling. The proposed method is hence described in Section~\ref{sec:method_description}, followed by a presentation of the validation results (Section~\ref{sec:results}); in addition to that conclusions and perspectives are presented in Section~\ref{sec:conclusion}.

\section{Method Description} \label{sec:method_description}

\subsection{Preliminaries}

This section presents a recap of the approach proposed in~\cite{mildenhall2020nerf}, which is considered in this article. NeRF represents the scene as a continuous volumetric field using an MLP. Its inputs are the position $\mathbf{x}_k$ and view direction $\mathbf{d}_k$ of a 3D point in space $\mathbf{p}_k$. Its outputs are the corresponding color $\mathbf{c}_k$ and density $\tau_k$:

\begin{equation}
[\mathbf{c}_k, \tau_k] = \mathrm{MLP} \left( \gamma( \mathbf{x}_k, \mathbf{d}_k); \theta \right).
\end{equation}

Here $\gamma(.)$ is an encoding function that maps positions and directions to a high-dimensional sine-cosine space:

\begin{equation}
\gamma(\mathbf{x}) = [\mathrm{sin}(\mathbf{x}), \mathrm{cos}(\mathbf{x}), ..., \mathrm{sin}(2^{L-1} \mathbf{x}), \mathrm{cos}(2^{L-1} \mathbf{x})]^T,
\end{equation}
with $L$ being a hyperparameter. The images are rendered by casting a ray $\mathbf{r}(t) = \mathbf{o} + t \mathbf{d}$ from the camera's center of projection $\mathbf{o}$ through each image pixel, with $t$ representing the distance along the ray. Therefore, the sample distances $t_{k} \in \mathbf{t}$ are drawn according to a sampling strategy along each ray between the near ($t_{n}$) and far ($t_{f}$) bounds ($\mathbf{t}$ is the vector of the samples). These resulting sampling distances are then used to compute the 3D positions of the input points, while the input direction is defined by the ray direction. After that, this set of sample points (position and direction) is inputted into the neural net to predict the corresponding colors and densities. The pixel values are finally computed using an approximate volume rendering integral \cite{Max1995OpticalMF}:

\begin{equation}
\mathbf{C(r; t, \theta)} = \sum_{k} w_{k} \mathbf{c}_{k},
\end{equation}
where
\begin{equation}
w_{k} = T_k \Big( 1 - \mathrm{exp} \big( -\tau_{k}(t_{k+1} - t_{k}) \big) \Big) \mathbf{c}_{k},
\end{equation}
and
\begin{equation}
T_k = \mathrm{exp}\left(-\sum_{k'<k} \tau_{k'}(t_{k'+1} - t_{k'})\right).
\end{equation}

To increase the sampling efficiency, NeRF trains a coarse and fine MLP, through minimization of the following loss:

\begin{equation}\label{eq:loss_photo}
\mathcal{L} = \sum_{\mathbf{r} \in R} \left( ||\mathbf{C}_{coarse} - \mathbf{C}^{*}_{coarse}||_{2}^{2} + ||\mathbf{C}_{fine} - \mathbf{C}^{*}_{fine}||_{2}^{2} \right).
\end{equation}

The parameters of $\mathbf{C}$ were omitted for readability. $\mathbf{C}$ and $\mathbf{C}^*$ are, respectively, the predicted and target pixel colors for the coarse and fine networks. $R$ represents the set of rays for training and $||.||_{2}$ is the L2 norm.

\subsection{Probabilistic Learning and Training}


This work proposes to leverage uncertainty information (e.g. uncertainty along viewing rays) to train neural radiance fields in a probabilistic manner. NeRF-based approaches tend to fail to produce consistent novel views when training data are overly uncertain. This is the case of vanilla NeRF which assumes absolute uncertainty along each viewing ray. NeRF overcomes this absolute uncertainty by using a large number of training views which allows to improve geometric consistency. One way to circumvent this drawback is to use additional probabilistic information during model training. 

\begin{figure}
	\centering
	\includegraphics[width=0.8\linewidth]{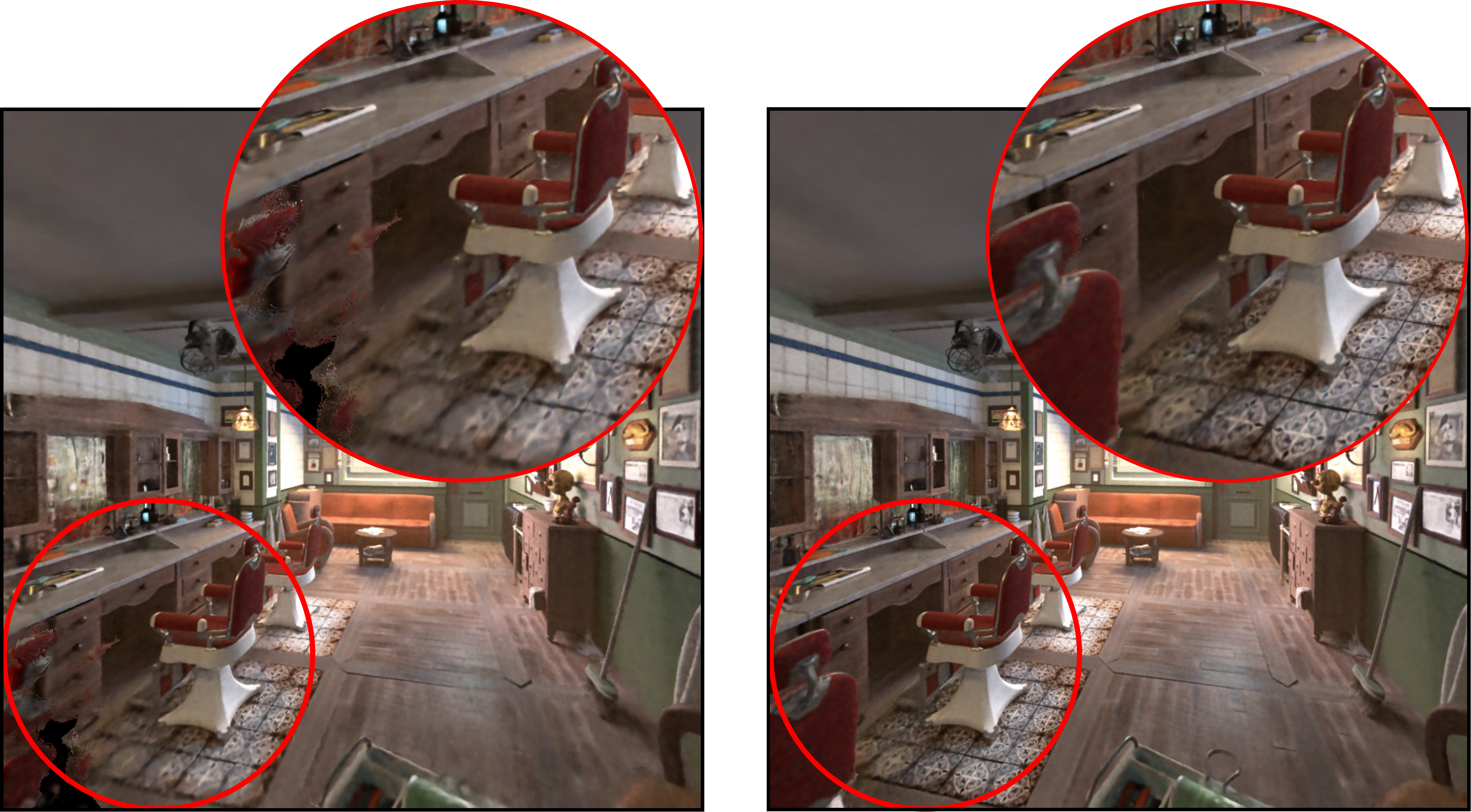}
	\caption{(left) image rendered using DONeRF trained with 60 images, (right) the same image rendered using DONeRF trained with 210 images. The method requires the full training set of images to learn a correct scene model.} 
	\label{fig:failure_example}
\end{figure}

In this paper, the case of depth uncertainty along the rays is considered because it is a primary source of uncertainty in NeRF architectures. Knowledge of depth uncertainty can be obtained from uncertain depth sensor measurements, uncertain correspondences between images, or even pose uncertainty. However, the uncertainty along rays is not explicitly modeled by the classic approach, which instead relies on the training process to recover it. As will be shown, such a strategy is indeed non-optimal, as it induces a need for additional data (views) to learn the scene representation. For example, consider Figure~\ref{fig:failure_example}, which shows an example of a rendered image using DONeRF \cite{neff2021donerf} for two cases (training with a subset of the training dataset and with the full training dataset).

\begin{figure}
	\centering
	\includegraphics[width=0.45\textwidth]{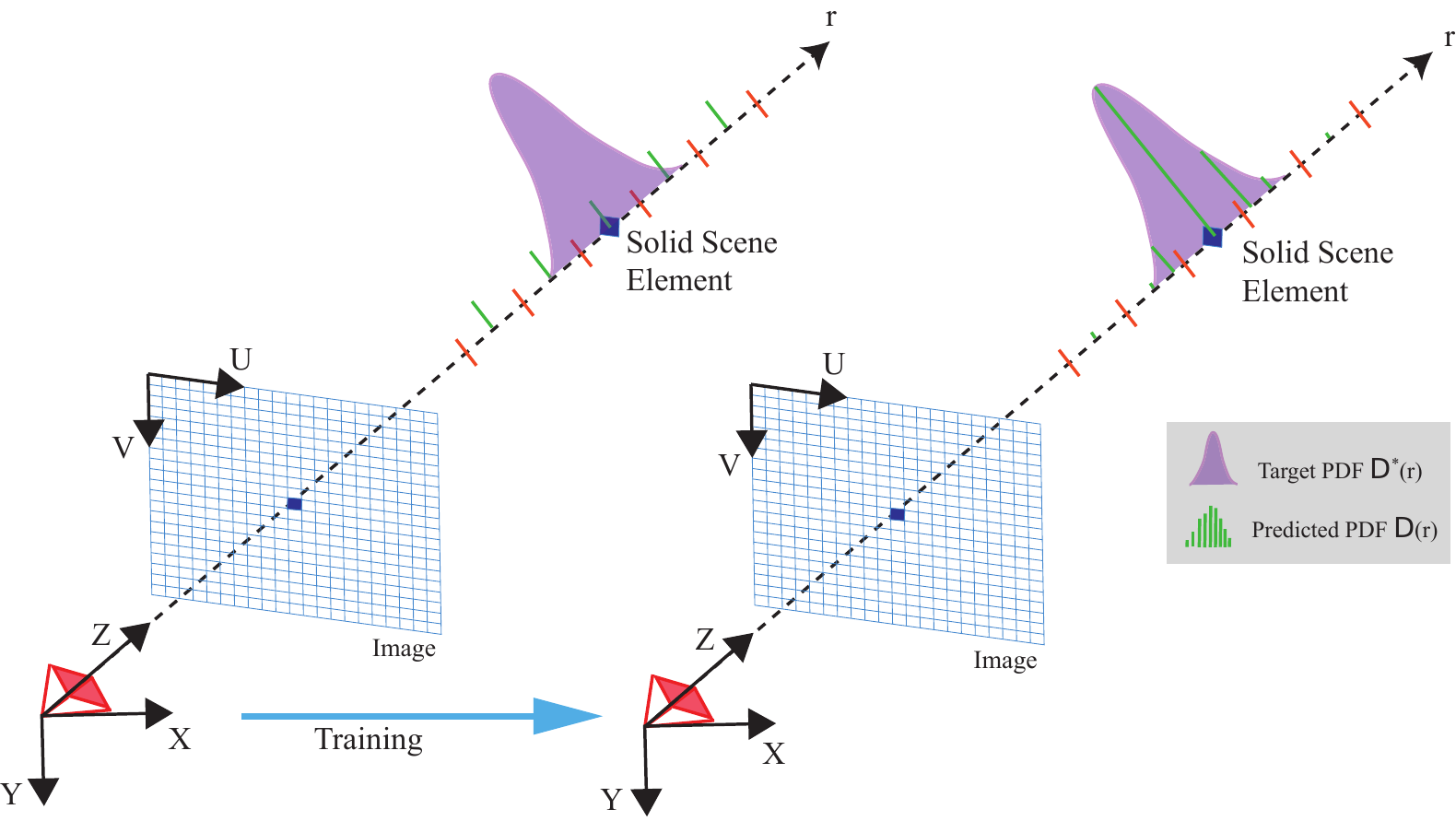}
	\caption{The proposed approach aims to supervise the learning of the depth PDF obtained from the network $\mathcal{D}(r)$ using a target PDF $\mathcal{D}^*(r)$. Left, the initial PDF in green has high uncertainty and right, after training the learnt PDF matches the target uncertainty.}
	\label{fig:depth_density_illustration}
\end{figure}

The proposition of this work is to integrate uncertainty into the training process and demonstrate its effectiveness through the example of depth uncertainty. This is achieved by fitting a depth PDF (Probability Density Function) obtained from the NeRF-based model $\mathfrak{D}(r)$ to a target ray PDF $\mathfrak{D}^*(r)$ obtained by various means. The former is computed by normalizing the weights $w_k$ along each ray \cite{mildenhall2020nerf} as in Equation (\ref{eq:depth_pdf_est}), while the latter is computed using a parametric PDF. The target PDF can be computed using available geometric information; such as depth sensor measurements $\mathbf{D}_i$, image correspondence/3D point uncertainty, camera pose uncertainty, and/or SFM algorithms (e.g. COLMAP~\cite{schoenberger2016sfm}). Figure~\ref{fig:depth_density_illustration} shows an illustration of the proposed representation of depth uncertainty. In the rest of this article, the target uncertainty will be modeled by a Gaussian PDF, centered around a depth value.

\begin{equation} \label{eq:depth_pdf_est}
\mathfrak{D}(r) = \frac{1}{\sum_{j} w_j}[w_0,..., w_N] \: / \: j = {0, 1, ..., N}. 
\end{equation}

$N$ and $[w_0,..., w_N]$ represent, respectively, the number of samples and the sample vector along the considered ray. One of the advantages regarding this approach is its ability to constrain depth uncertainty (inherent to depth measurements or 3D points) through a target uncertainty distribution. It is therefore possible to supervise the learned depth in accordance with the uncertainty sources (such as the uncertainty on the 3D point position or the camera position).
Equation (\ref{eq:color_depth_density_loss}) shows the proposed loss function, which takes into account depth uncertainty within the training process.

\begin{equation} \label{eq:color_depth_density_loss}
\mathcal{L} = K_{color} \, \sum_{r \in R} \mathcal{L}_{color} + K_{density} \, \mathcal{L}_{density} + K_{depth} \, \mathcal{L}_{depth},
\end{equation}
with
\begin{equation}
\mathcal{L}_{color} = ||\mathbf{C}_{coarse} - \mathbf{C}^{*}_{coarse}||_{1} + ||\mathbf{C}_{fine} - \mathbf{C}^{*}_{fine}||_{1},
\end{equation}

\begin{equation}
\mathcal{L}_{density} = ||\mathfrak{D}_{coarse} - \mathfrak{D}^*_{coarse}||_{1} + ||\mathfrak{D}_{fine} - \mathfrak{D}^*_{fine}||_{1},
\end{equation}
and

\begin{equation}
\mathcal{L}_{depth} = ||\mathbf{D}_i - \mathbf{D}_i^*||_{1}.
\end{equation}

\begin{figure}[h]
	\centering
	\includegraphics[width=0.32\textwidth]{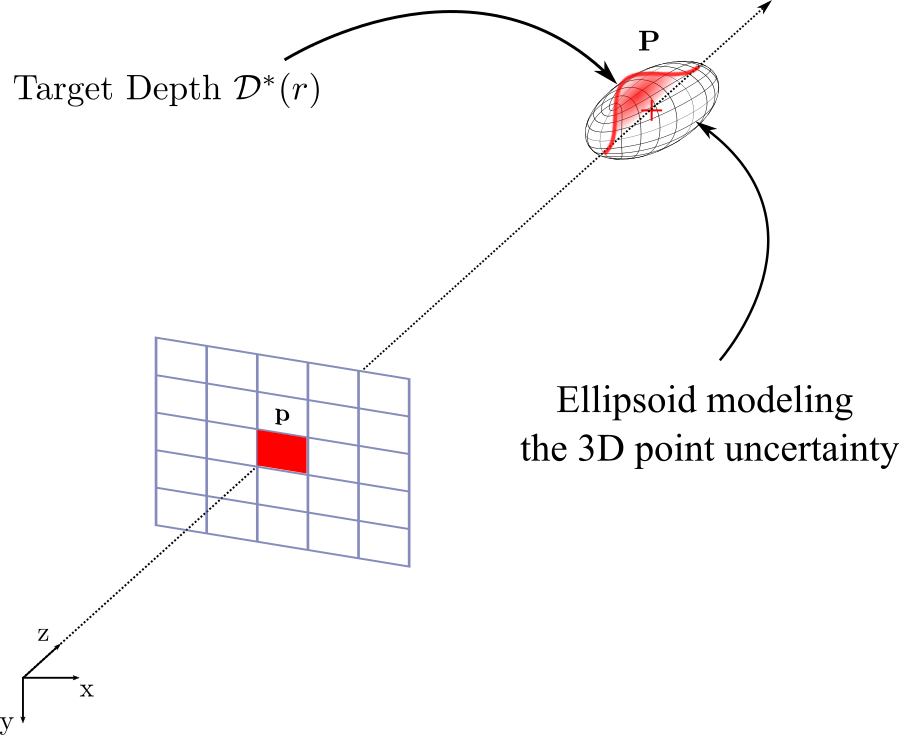}
	\caption{Target depth PDF $\mathcal{D}^*(r)$ obtained from the uncertain estimate of a 3D point. In this configuration, $\mathcal{D}^*(r)$ allows to take into account the pose uncertainty of the camera along the $z$ axis.}
	\label{fig:3d_point_pnerf}
\end{figure}

The depth PDF parameters were omitted for readability. $||.||_1$ is the L1 norm. This was preferred over the L2 norm used by NeRF due to the higher error values it produces when the error is smaller than 1 (which is the case for the depth densities). Moreover, it exhibits increased robustness toward outliers.  $K_{color}$, $K_{density}$, and $K_{depth}$ are the gains applied to the terms. The depth $\mathbf{D_i}$ is computed as in Equation (\ref{eq:depth}), while $\mathbf{D_i}^*$ is a reference depth that can be obtained from a depth measurement or a point cloud as shown in Figure \ref{fig:3d_point_pnerf}. It is worth noting that the Gaussian PDF represented in Figure \ref{fig:3d_point_pnerf} takes into account the uncertainty of the 3D point in addition to the one of the camera pose along the $z$ axis.   

\begin{equation} \label{eq:depth}
\mathbf{D}_i = \sum_{k} w_k \mathbf{t}_k.
\end{equation}

The depth term in this loss function aims to bias the learning process towards the most probable depth (the mean value), while taking into account its variance through the density 
term.

\section{Results} \label{sec:results}

In this section, the effect of the different cost function terms is studied first, showing the benefit of the proposed loss. This is followed by a comparison with state-of-the-art methods on both synthetic and real data. The models considered here were trained on an Nvidia GeForce RTX 3090 GPU with 24 Gb RAM. In addition to that, the implementation of Pytorch Lightning of NeRF\footnote {https://github.com/kwea123/nerf\_pl/} was used to develop the proposed approach. Noting that the target Gaussian considered in these experiments is a unit Gaussian PDF centered around the target depth. 


\subsection{Ablation study}

This part presents a comparison between NeRF models trained with the different terms of the loss described in Equation (\ref{eq:color_depth_density_loss}). The first model is trained with pure photometric loss (Equation (\ref{eq:loss_photo})) as in the original NeRF. The loss of the second model, for its part, contains only the color and density terms (modeling the uncertainty). The loss of the third model contains all the terms of Equation (\ref{eq:color_depth_density_loss}). Training is done using a stereo image pair that was generated from Blender using the Sponza virtual scene\footnote {https://www.intel.com/content/www/us/en/developer/topic-technology/graphics-research/samples.html}. In addition, two additional stereo pairs were generated for validation and testing. The objective here is to provide insights about the influence of the loss terms on the quality of the learnt representation.

\begin{figure}
	\centering
	\includegraphics[width=\linewidth]{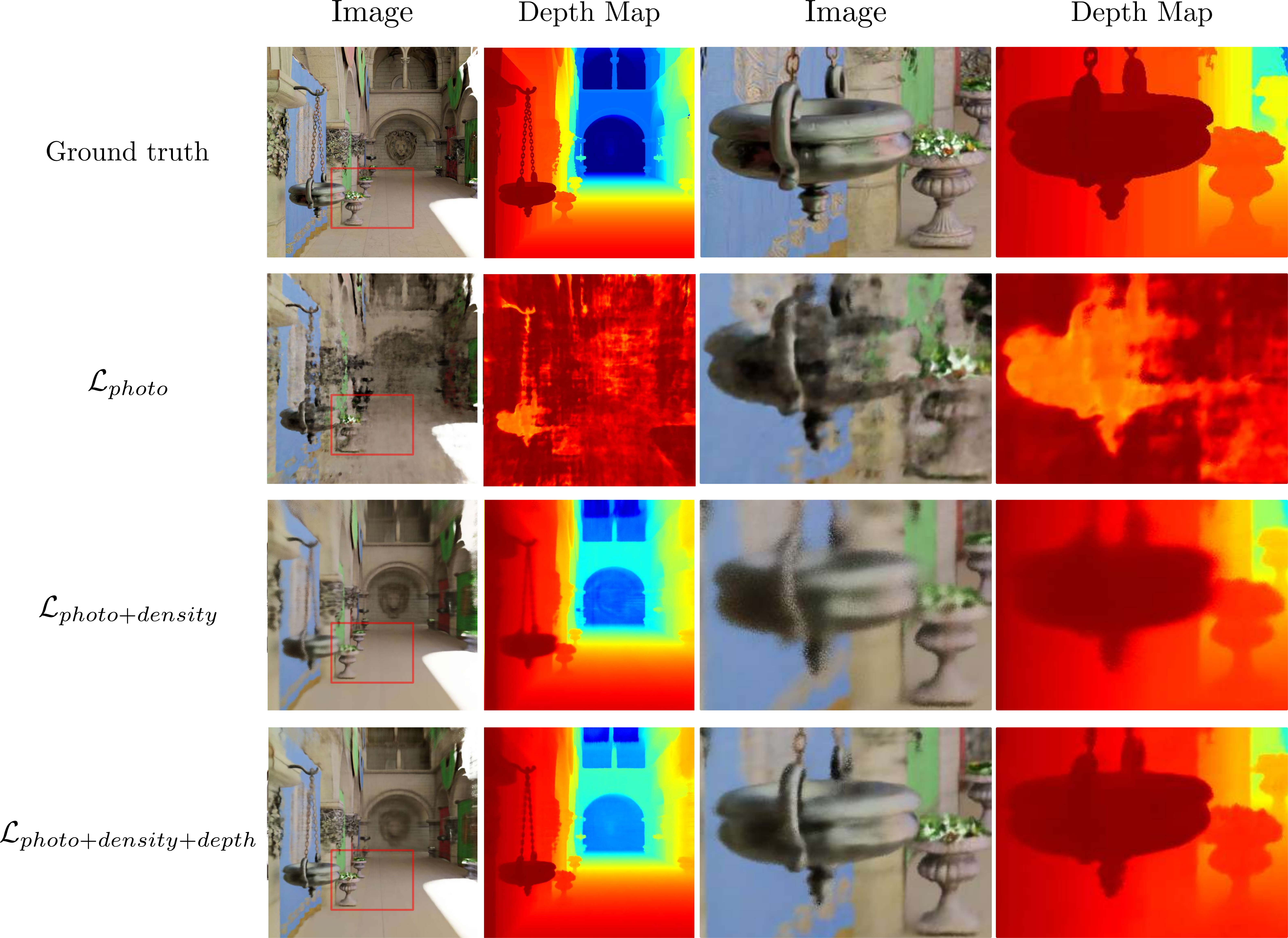}
	\caption{Qualitative results for the ablation study. The introduction of depth information allows to better model the scene. Adding only the depth uncertainty term is not sufficient to allow the model to learn sharp scene elements, while biasing it towards the mean depth allows for a better scene modeling.}
	\label{fig:ablation_qualitative}
\end{figure}

Table \ref{table:ablation_quantitative} shows the PSNR values (Peak Signal-to-Noise Ratio) obtained with the different loss terms. It can be seen that considering only the photometric term provides the lowest PSNR value, as the model cannot learn a geometrically consistent scene representation (Figure \ref{fig:ablation_qualitative}). Integrating depth uncertainty through the addition of the density term to the loss allows to synthesize images with better quality. This is notably due to the fact that the learned scene geometry is consistent, allowing for a better generalization. However, sharp elements are not correctly learned, as shown in Figure \ref{fig:ablation_qualitative}. The term related to depth uncertainty, in fact, allows the model to learn a distribution of the depth corresponding to the image pixels, inducing a smoothing of the predicted depth. Therefore, the edges are poorly modeled. This aspect is compensated by adding to the depth term the loss, which forces the model to predict depth values that tend towards mean depth. 

\begin{table}
	\centering
	\begin{tabular}{c | c}
		\hline
		Terms & PSNR $\uparrow$ \\
		\hline
		$\mathcal{L}_{photo}$ & 14.44 \\
		
		$\mathcal{L}_{photo + density}$ & 20.66 \\
		
		$\mathcal{L}_{photo + density + depth}$ & \textbf{20.85} \\
		
	\end{tabular}
	\caption{PSNR values for the considered loss functions. Adding the depth uncertainty term to the loss allows for a better modeling of the scene, but does not allow to model sharp elements. Adding the depth term allows to achieve a better modeling of sharp scene elements.}
	\label{table:ablation_quantitative}
\end{table}

\subsection{Validation on Synthetic Data}

The dataset provided by \cite{neff2021donerf} was considered to validate the proposed approach. More specifically, the \textit{Classroom} and \textit{Barbershop} sequences that represent indoor synthetic scenes were considered as the proposed approach aims at modeling scenes with finite depth. The metrics considered here are the PSNR and the SSIM (Structural SIMilarity) that allow to measure the level of corruption of the rendered images and the perceptual similarity to the test images, respectively. Note that the model trained with the proposed loss function is referred to as P-NeRF, which renders $200\times 200$ images to speed up training. The models were trained with ADAM \cite{kingma2014adam} for $200$ epochs with a batch size of $2048$ and a learning rate of $0.0005$. The total number of evaluations is $256$ per iteration. In addition, the checkpoints used for the test are those with the lowest PSNR at the test time.

P-NeRF is compared with the DONeRF approach \cite{neff2021donerf} that uses an oracle depth network to produce a sampling region for the NeRF model to render $400\times 400$ images. The publicly available code\footnote {https://github.com/facebookresearch/DONERF} was used for this evaluation. Furthermore, the models were trained with $20$, $60$, $140$, and full sequence lengths. This validation allowed to study the effect of the size of the training set on the generalization. 

Tables \ref{table:synthetic_data_partial_quantitative_barbershop} and \ref{table:synthetic_data_partial_quantitative_classroom} show the PSNR and SSIM values on the complete test sets of the Barbershop and Classroom sequences, for both P-NeRF and DONeRF using different training set sizes (20, 60, and 140 images). It can be seen that P-NeRF renders better image quality at test time compared to DONeRF, when the training set contains 20, 60, and 140 randomly selected images. This supports the claim that the integration of the depth information in a probabilistic manner allows the proposed P-NeRF to learn a better model of the scene, resulting in better-rendering images.

\begin{table}
	\centering
	\begin{tabular}{c | c | c | c}
		\hline
		Training Images  & Method & PSNR $\uparrow$ & SSIM $\uparrow$  \\
		
		\Xhline{1pt}
		\multirow{2}{5em}{20 Images}  & DONeRF & 26.75  & 0.87 \\
		
		& P-NeRF & \textbf{27.23} & 0.87 \\
		\Xhline{0.7pt}
		\multirow{2}{5em}{60 Images} & DONeRF & 27.68  & 0.87 \\
		
		& P-NeRF & \textbf{31.30} & \textbf{0.93} \\
		\Xhline{0.7pt}
		\multirow{2}{5em}{140 Images} & DONeRF & 29.07 & 0.89 \\
		
		& P-NeRF & \textbf{32.92} & \textbf{0.95}  \\
		\Xhline{0.7pt}
	\end{tabular}
	\caption{Quantitative results for the DONeRF and the proposed P-NeRF using different training sets from the Barbershop sequence. The proposed approach permits to predict images with a better quality than the DONeRF, when the training set is limited.}
	\label{table:synthetic_data_partial_quantitative_barbershop}
\end{table}

\begin{table}
	\centering
	\begin{tabular}{c | c | c | c}
		\hline
		Training Images  & Method & PSNR $\uparrow$ & SSIM $\uparrow$  \\
		
		\Xhline{1pt}
		\multirow{2}{5em}{20 Images}  & DONeRF & 28.83 & 0.90 \\
		
		& P-NeRF & \textbf{29.85} & 0.90 \\
		\Xhline{0.7pt}
		\multirow{2}{5em}{60 Images} & DONeRF & 29.83  & 0.88  \\
		
		& P-NeRF & \textbf{31.88} & \textbf{0.92} \\
		\Xhline{0.7pt}
		\multirow{2}{5em}{140 Images} & DONeRF & 34.53 & 0.94 \\
		
		& P-NeRF & \textbf{35.54} & \textbf{0.97}  \\
		\Xhline{0.7pt}
	\end{tabular}
	\caption{Quantitative results for the DONeRF and the proposed P-NeRF using different training sets from the Classroom sequence. The proposed approach permits to predict images with a better quality than the DONeRF, when the training set is limited.}
	\label{table:synthetic_data_partial_quantitative_classroom}
\end{table}

In this part of the validation, the full training sets (210 views) are considered, and the proposed P-NeRF is compared with methods from state-of-the-art. The results of this comparison are shown in Table \ref{table:synthetic_data_full_quantitative}, where the PSNR values of the baseline methods were taken from Table 2 of \cite{neff2021donerf}. Therefore, the SSIM metric is not considered here as it was not presented in \cite{neff2021donerf}. The considered state-of-the-art baseline methods are listed below: 
\begin{itemize}
	\item \textbf{DONeRF} \cite{neff2021donerf}:  the variant (in terms of sampling strategy and use of ground truth depth) that provides the best image quality in each case is considered here.
	\item \textbf{NeRF} \cite{mildenhall2020nerf}: the results shown here were obtained using a NeRF PyTorch implementation (https://github.com/yenchenlin/nerf-pytorch) with a total of $256$ network evaluations per iteration. The mention (log + warp) refers to the sampling strategy proposed by \cite{neff2021donerf} that was applied to the original NeRF.
	\item \textbf{NeX (Neural Basis Expansion)} \cite{Wizadwongsa2021NeX}: the implementation provided by the authors (https://github.com/nex-mpi/nex-code/) was considered.
	\item \textbf{LLFF (Local Light Field Fusion)} \cite{mildenhall2019llff}: the open source code available at (https://github.com/Fyusion/LLFF) was used for this evaluation.
\end{itemize}

It is possible to see that considering the full training set induces an increase in the quality of the images rendered by DONeRF and P-NeRF. The level of quality reached by the proposed approach, however, is the highest, as all the baseline methods produce lower image quality at test time. This result shows that P-NeRF is better able to generalize to novel view rendering.

\begin{figure}
	\centering
	\includegraphics[width=0.95\linewidth]{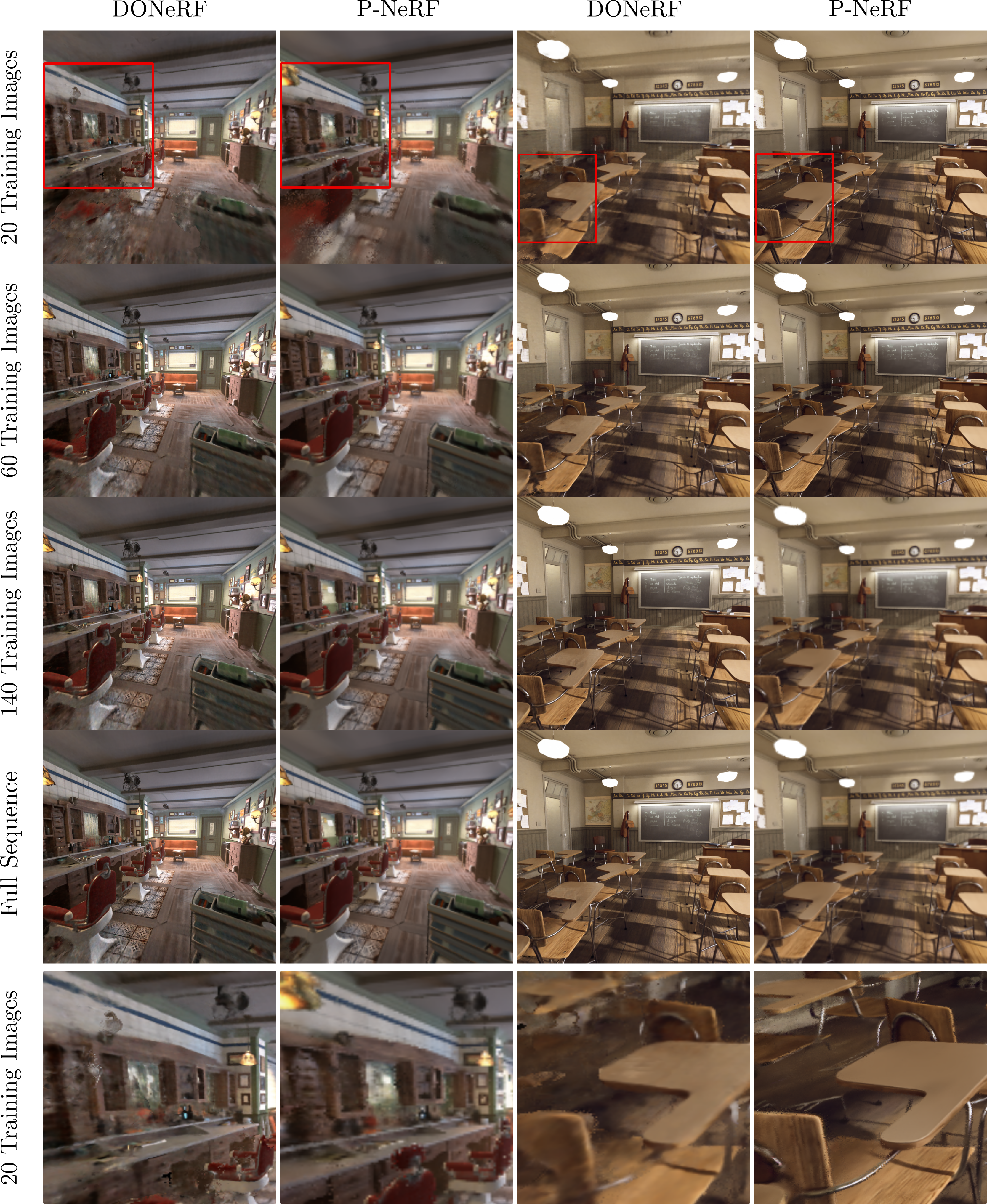}
	\caption{Example images rendered using P-NeRF and DONeRF. It is possible to notice that more details are present in the images rendered by P-NeRF, and that scene elements (chair, lamp) are dropped by DO-NeRF while being modeled by P-NeRF.}
	\label{fig:synthetic_data_full_quantitative}
\end{figure}

\begin{table}
	\centering
	\begin{tabular}{c | c | c | c}
		\hline
		Method & Barbershop & Classroom & Mean \\
		\Xhline{1.0pt}
		P-NeRF & \textbf{34.05} & \textbf{36.74} & \textbf{35.40} \\
		
		DONeRF & 32.80 & 36.27 & 34.54 \\
		
		NeRF & 33.63 & 34.02 & 33.83\\
		
		NeRF (log + warp) & 33.60 & 35.19 & 34.40 \\
		
		NeX & 22.98 & 30.34 & 26.66\\
		
		LLFF & 24.13 & 24.87 & 24.50\\
		
		\hline
		
	\end{tabular}
	\caption{Values of the PSNR for the considered methods trained on the full training sets (210 images). The values of the baseline methods are obtained from Table 2 in \cite{neff2021donerf}. The proposed approach predicts images with a better quality than the DONeRF, when the training set is limited.}
	\label{table:synthetic_data_full_quantitative}
\end{table}


The result of a qualitative comparison between P-NeRF and DONeRF is presented in Figure \ref{fig:synthetic_data_full_quantitative}. It shows that the proposed P-NeRF is capable of learning a consistent representation of the scenes, permitting to render images with a better quality in comparison with DONeRF. This difference is especially apparent when the training data are limited (20 training views); as this is the case for the Barbershop sequence, where P-NeRF is able to learn to represent scene elements that are dropped by DONeRF.

The results of the validation on synthetic data show that the proposed P-NeRF outperforms the baseline state-of-art methods. This is notably due to the fact that P-NeRF learns a consistent scene representation that permits to achieve a better generalization,  and this even when the training data is limited.


\subsection{Validation on Real Data}
The results presented in this part are related to the real dataset provided by \cite{mildenhall2020nerf}. In order to learn the scenes models, the poses of the training frames and a sparse point cloud are first computed using COLMAP \cite{schoenberger2016sfm}. The sparse set of 3D points is then used to integrate depth information into the training process in the following way. The pixels corresponding to the sparse set of estimated 3D points are obtained through projection. Therefore, it is possible to consider for each of these pixels the loss described in Equation (\ref{eq:color_depth_density_loss}), where the corresponding target depth PDF is an approximation of the intersection between the ray passing by the pixel and the ellipsoid representing the uncertainty of the projected 3D point (Figure \ref{fig:3d_point_pnerf}). This allows to account for the uncertainty of the pose that propagates into the camera's $z$ axis. For its part, the corresponding depth is the distance between the camera and the 3D point. In addition to that, pixels with high reprojection error are down-weighted similarly to \cite{kangle2021dsnerf} in order to favor the most precise depth estimates during training. For the remaining set of pixels, only the photometric term is considered. 

The comparison is made between the proposed P-NeRF and DS-NeRF \cite{kangle2021dsnerf}. The methodology of \cite{kangle2021dsnerf} was followed to select the test images with indexes {$0$, $8$, $16$, ...} until the end of the sequences. The training set for its part contains $10$ randomly selected images. The number of network evaluations per iteration is 128.

\begin{figure}
	\centering
	\includegraphics[width=0.7\linewidth]{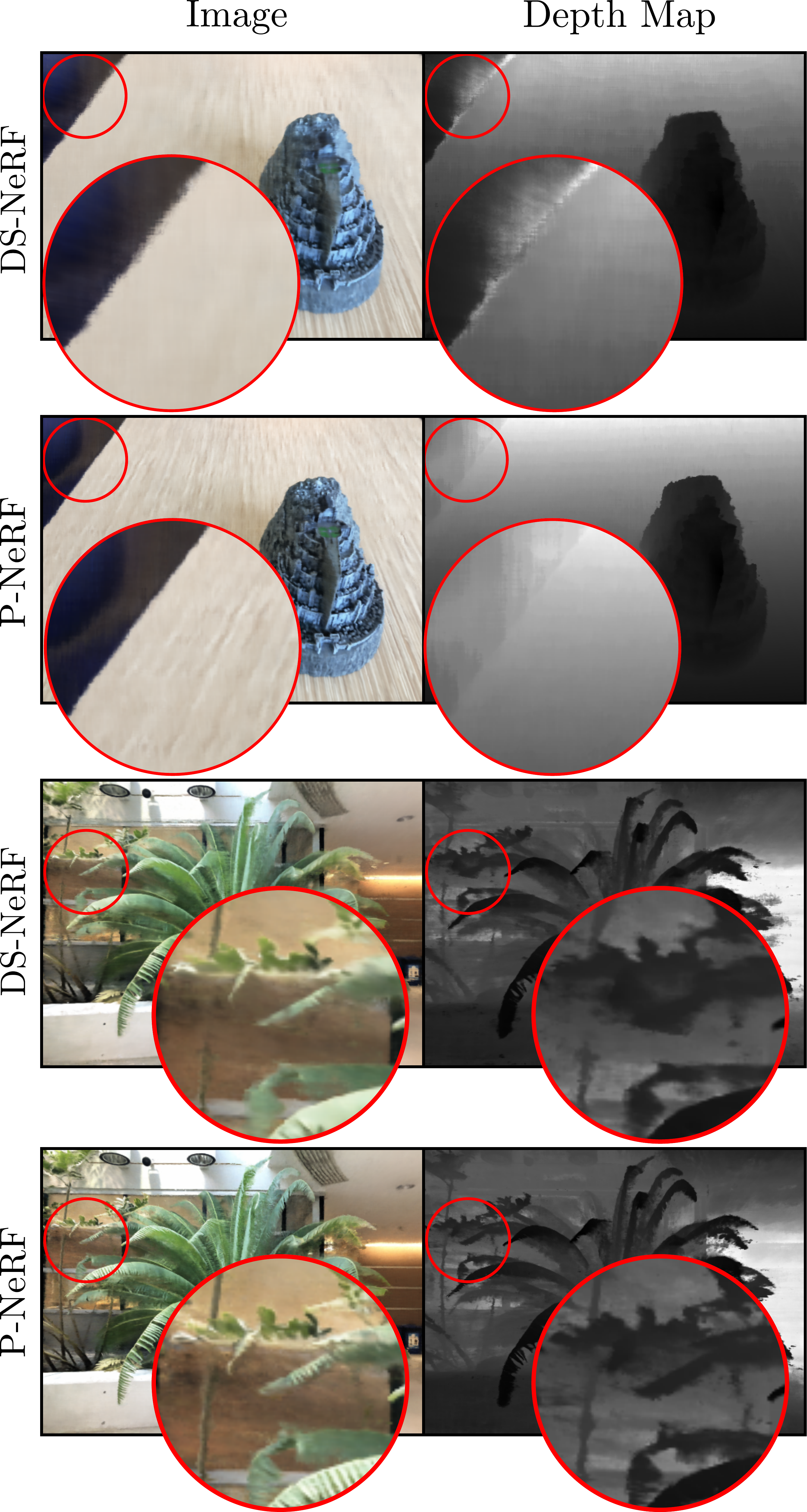}
	\caption{Rendered images and depth maps using DS-NeRF \cite{kangle2021dsnerf} and P-NeRF with a training set of 10 images. The proposed approach is able to learn more details from the scene.}
	\label{fig:qualitative_real_data}
\end{figure}

Table \ref{table:quantitatve_real} shows the PSNR and SSIM values reached in the test set for both P-NeRF and DS-NeRF. The proposed method produces better image quality on the test set for both the Fern and Fortress sequences and has the best mean metrics in the test set. This difference in the quality of the rendered images results from the fact that the proposed approach is capable of learning the details of the scene, which are not modeled by DS-NeRF, as shown in Figure \ref{fig:qualitative_real_data}.

\begin{table}[h]
	
	\centering
	\begin{tabular}{ c | c | c | c | c}
		\hline
		\multirow{2}{5em}{Sequence} & \multicolumn{2}{ c | }{DS-NeRF} & \multicolumn{2}{ c }{P-NeRF} \\
		\cline{2-5}
		& PSNR $\uparrow$ & SSIM $\uparrow$ & PSNR $\uparrow$ & SSIM $\uparrow$ \\
		\Xhline{1pt}
		Fern & 24.27   & 0.74  & \textbf{25.53} & \textbf{0.80} \\
		
		Fortress & 23.48  & 0.62 & \textbf{24.13}  & \textbf{0.70} \\
		
		Flower & \textbf{23.08}  & \textbf{0.68} & 22.45  & 0.66 \\
		\hline
		
	\end{tabular}
	\caption{Quantitative results for the DS-NeRF and the proposed P-NeRF for different training sets. The proposed approach produces better image quality at test time.}
	\label{table:quantitatve_real}
\end{table}

\section{Conclusions and Perspectives} \label{sec:conclusion}
This paper has proposed a novel method to account for uncertainty in the learning process of neural scene representations and has subsequently made a first step in providing a probabilistic representation that is fundamental for robotics applications. A case study was made based on depth uncertainty to supervise the depth PDF obtained from the NeRF using the available depth information to learn a geometrically consistent scene representation. This explicit introduction of uncertainty supervision into the learning process allowed to take into account different sources of depth uncertainty (depth maps, point clouds, and camera pose). As shown in the results, such probabilistic learning allows to enhance the learning process and permits to reach better image rendering quality compared to state-of-the-art methods, which also use depth information.

One perspective of this approach is to consider different uncertainty model sources in the learning process in order to further improve the quality of the scene representation. For example, the proposed approach could potentially account for probabilistic sources of information such as uncertain stereo matching, uncertain depth prediction, uncertain semantic segmentation, camera extrinsic and intrinsic uncertainty, etc.  Further works will involve adapting real-time NeRF approaches~\cite{mueller2022instant} to take into account this probabilistic model for application to a real-world robotic navigation task.


\section*{ACKNOWLEDGMENT}

This work was funded by the EU H2020 MEMEX research project under grant
agreement No. 870743. Parts of the experiments that led to this article were performed using HPC resources from GENCI-IDRIS (Grant 2022-011012578).

\bibliographystyle{plain}
\bibliography{references}


\end{document}